\documentclass[11pt]{article}
\usepackage[margin=1in]{geometry}
\usepackage{booktabs}
\usepackage{amsmath,amssymb}
\usepackage{graphicx}
\usepackage{xcolor}
\usepackage{hyperref}
\usepackage{authblk}
\usepackage{pifont}

\title{\textbf{Nystr\"{o}m Attention Matches Full Attention\\for Cross-Sectional Stock Prediction}}

\author{Kunhan Guo}
\affil{The University of Hong Kong \\ \texttt{aguo0521@connect.hku.hk}}

\date{}

\begin{document}
\maketitle

\begin{abstract}
MASTER's inter-stock multi-head attention---the module responsible for modeling cross-sectional stock relationships---accounts for 42.5\% of model parameters and 25\% of predictive value. We systematically decompose this module and uncover a surprising structure: the learned attention is near-uniform (perplexity 278/300), yet forcing exact uniformity eliminates all cross-sectional discrimination. Spectral analysis resolves this paradox: the deviation from uniformity is \emph{low-rank} (effective rank $\sim$65, top-10 modes capture 96.5\% of energy), explaining why sparse approximations consistently fail while Nystr\"{o}m low-rank attention ($m=32$ landmarks) matches full $O(N^2)$ attention at $O(mN)$ cost---certified equivalent via TOST at both $N=300$ (5 seeds, Rank~IC $p=0.003$) and $N=800$ (10 seeds, Rank~IC $p=0.034$). Additional findings include: (i)~attention anti-correlates with return similarity (Spearman $\rho = -0.614$; on the industry-labeled subset, $-0.645$ unconditionally and $-0.627$ after controlling for industry, beta, and volatility), suggesting complementarity-seeking rather than correlation mining; (ii)~all graph-based alternatives degrade performance, with hard masking worse than complete module removal; and (iii)~at $N \approx 3{,}500$ with adapted architectures, no cross-stock module (GCN, Nystr\"{o}m, or MASTER-style pipeline) significantly outperforms a per-stock LSTM baseline ($n=4$ seeds), indicating that the benefits observed at smaller scales do not trivially transfer. These results establish that the inter-stock attention's value resides in a compressible, dynamic, near-global redistribution that rewards low-rank approximation but resists sparsification.
\end{abstract}

\section{Introduction}

Cross-sectional stock prediction---ranking which stocks will outperform within a universe---is a core task in quantitative finance. Recent Transformer-based models, particularly MASTER (Market-Guided Stock Transformer)~\cite{master}, have become standard baselines by applying multi-head attention \emph{across stocks} at each timestep. This inter-stock attention module (Step~\ding{194} in MASTER's five-stage pipeline) claims to capture ``momentary and cross-time stock correlations,'' and subsequent work has sought to enhance it with explicit graph structures drawn from industry membership, supply chains, or return correlations~\cite{thgnn,tgns,clgnn,ci-sthpan,finmamba,lsr-igru,s3g}.

But what does this module actually learn? And can it be made cheaper? We address both questions through a systematic decomposition combining statistical diagnostics, controlled ablations, spectral analysis, and efficient-attention experiments. Our findings challenge several prevailing assumptions:

\begin{enumerate}
\item \textbf{The attention is near-uniform but not uniform.} Per-stock entropy is within 1.2\% of the theoretical maximum, yet forcing exact uniformity eliminates all cross-sectional discrimination. The deviation from uniformity contributes only 1.3\% of output energy but is the \emph{sole} source of cross-sectional variance.

\item \textbf{The deviation structure is low-rank.} Spectral analysis of the deviation matrix $D = A - \frac{1}{N}\mathbf{1}\mathbf{1}^\top$ reveals an effective rank of $\sim$65 (out of 300 stocks), with the top-10 singular values capturing 96.5\% of the Frobenius norm. This directly predicts that low-rank attention should succeed while sparse attention should fail.

\item \textbf{Nystr\"{o}m attention matches full attention at $O(mN)$.} Replacing full $O(N^2)$ attention with Nystr\"{o}m approximation using just $m=32$ landmarks produces equivalent Rank~IC (TOST $p=0.003$ within $\pm 0.005$) and equivalent IC within $\pm 0.008$ (TOST $p=0.028$) across 5 seeds, with \emph{lower} seed-to-seed variance.

\item \textbf{Sparsification consistently fails.} Graph-masked, top-$K$, and deviation-thresholded variants all degrade performance. Hard masking to correlation-based neighbors is worse than removing the module entirely. A 2$\times$2 factorial reveals that IC scales with neighbor \emph{count} while Rank~IC scales with weight \emph{sharpening}---requirements that are mutually exclusive under sparsity.

\item \textbf{Large-$N$ transferability is limited.} At $N \approx 3{,}500$ (full A-share market) with adapted architectures, no cross-stock module significantly outperforms a per-stock LSTM baseline across multiple seeds ($n=4$). The benefits documented at $N=300$--800 do not trivially transfer to larger, noisier universes---a finding that itself required multi-seed validation to establish, as single-seed results suggested a spurious trade-off.
\end{enumerate}

Together, these findings establish that MASTER's inter-stock attention implements a \emph{compressible, dynamic, complementarity-seeking} redistribution of cross-sectional information. Its value is real but resides in a low-rank structure that rewards Nystr\"{o}m approximation and resists sparsification---a distinction with direct implications for scaling stock Transformers, tempered by the observation that scaling benefits are not guaranteed at larger $N$.

\section{Background: MASTER Architecture}
\label{sec:background}

MASTER~\cite{master} processes $N$ stocks over $T$ lookback timesteps through five stages. \textbf{Step~\ding{192}} applies market-guided gating, using market index features to dynamically rescale each stock's input. \textbf{Step~\ding{193}} performs intra-stock temporal attention---a Transformer encoder operating independently per stock across $T$ timesteps. \textbf{Step~\ding{194}} applies inter-stock multi-head attention at each timestep:
\begin{equation}
z_{u,t} = \sum_{v=1}^{N} \alpha_{uv}^{(t)} V_v^{(t)}, \quad \alpha_{uv}^{(t)} = \mathrm{softmax}\!\left(\frac{Q_u^{(t)} K_v^{(t)\top}}{\sqrt{d}}\right)
\label{eq:attention}
\end{equation}
This module contains 329,216 parameters (42.5\% of the model), including QKV projections, a feed-forward network, and LayerNorm. \textbf{Step~\ding{195}} collapses the temporal dimension via attention (the last timestep queries all others), and \textbf{Step~\ding{196}} applies a linear prediction head.

\paragraph{Why focus on Step~\ding{194}?} We ablate each step individually (Table~\ref{tab:ablation_full}). Step~\ding{194} contributes the largest IC drop ($-$25.3\%) and is the only module enabling cross-stock information flow. Notably, removing Step~\ding{195} (temporal aggregation) \emph{increases} Rank~IC by 5.4\%, foreshadowing the aggregation-hurts-ranking phenomenon we document at scale in Section~\ref{sec:scale}.

\begin{table}[h]
\centering
\caption{Full pipeline ablation (CSI300, seed 0). Step~\ding{194} is the dominant value contributor. $\Delta$IC computed from unrounded values; displayed IC is rounded to 4 decimal places.}
\label{tab:ablation_full}
\begin{tabular}{lcccc}
\toprule
\textbf{Configuration} & \textbf{IC} & \textbf{Rank IC} & \textbf{$\Delta$IC} & \textbf{Params} \\
\midrule
Full MASTER & 0.0646 & 0.0685 & --- & 775,041 \\
No Step~\ding{192} (gating) & 0.0622 & 0.0673 & $-$3.6\% & 764,929 \\
No Step~\ding{193} (temporal attn) & 0.0612 & 0.0671 & $-$5.2\% & 445,825 \\
No Step~\ding{194} (inter-stock attn) & 0.0482 & 0.0514 & $-$25.3\% & 445,825 \\
No Step~\ding{195} (temporal agg.) & 0.0605 & \textbf{0.0722} & $-$6.3\% & 709,505 \\
\bottomrule
\end{tabular}
\end{table}

\section{Experimental Setup}
\label{sec:setup}

\paragraph{Data.} Our primary experiments use the MASTER opensource CSI300 dataset~\cite{master}: $\sim$300 A-share stocks with 222-dimensional features (158 Alpha158 technical factors, 63 market features, 1 label), following the original train/validation/test split (619 test days). For scale validation, we use the full A-share market via Qlib ($\sim$3,486 stocks per day on average), with 17 features computed from raw OHLCV, training on 2010--2017 and testing on 2019--2020 (383 days).

\paragraph{Metrics.} We report IC (Pearson correlation between predicted and actual 4-day forward returns), Rank~IC (Spearman correlation, measuring ordering accuracy), and their information ratios (ICIR = mean/std, measuring consistency). The label is defined as $\text{close}(t{+}5)/\text{close}(t{+}1) - 1$, a strictly future return with a one-day gap, following the MASTER opensource configuration. Long-short portfolio Sharpe ratios are reported for scale experiments.

\paragraph{Protocol.} All CSI300 experiments use MASTER's original loss-threshold stopping (train loss $\leq 0.95$) with Adam optimization. All Step~\ding{194} variants are implemented as subclasses without modifying the original codebase, ensuring identical treatment of Steps~\ding{192}, \ding{193}, \ding{195}, and \ding{196}. Single-seed (seed 0) results are used for mechanistic decomposition where relative ordering across many controlled variants is the focus; all equivalence claims are validated across 5--10 seeds with proper TOST testing.

\section{Attention Diagnostic}
\label{sec:diagnostic}

We extract inter-stock attention weights from a trained MASTER model across all 619 test days and characterize their statistical properties.

\begin{figure}[h]
\centering
\includegraphics[width=0.9\linewidth]{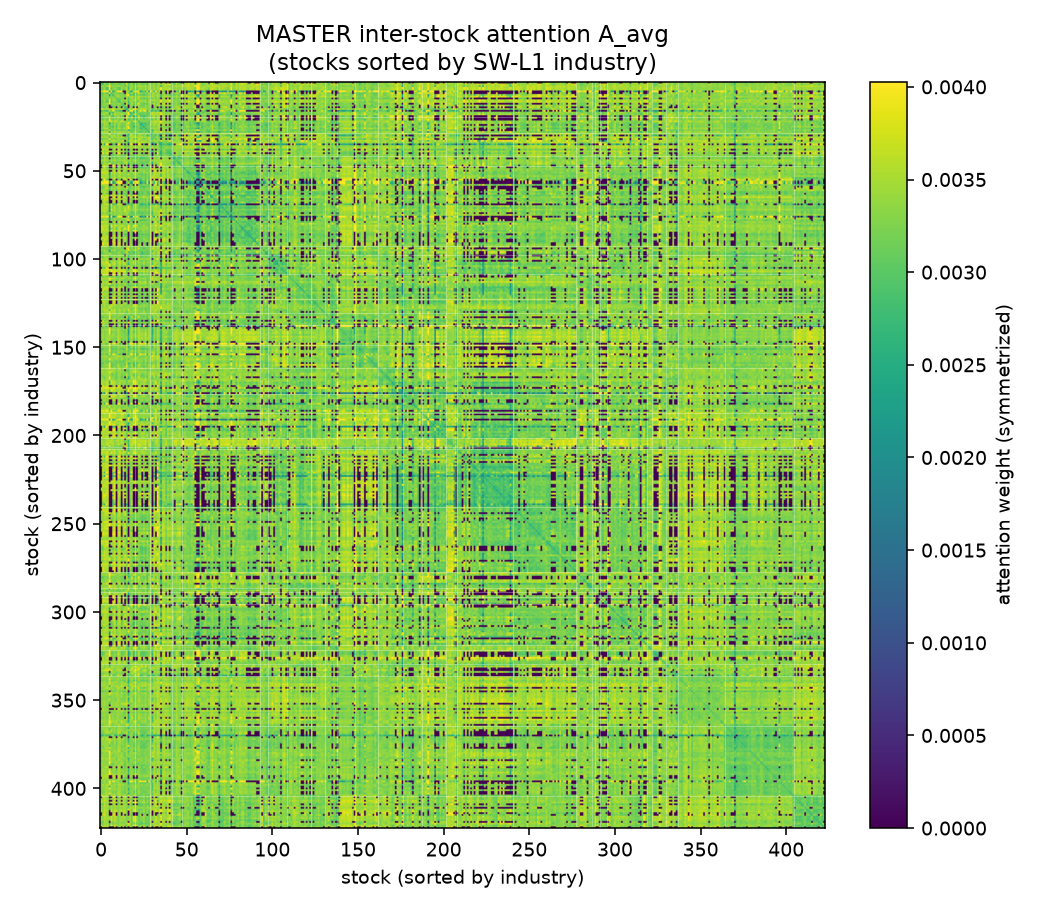}
\caption{Time-averaged attention matrix (centered at $1/N$). Blue = below-average, red = above-average. Stocks sorted by Shenwan L1 industry; no block-diagonal structure. The matrix shows 433 unique stocks appearing across test days; on any given day $\sim$300 are active.}
\label{fig:heatmap}
\end{figure}

\paragraph{Near-uniform distribution.} The mean per-row attention entropy is 5.63 nats, compared to $\ln(300) = 5.70$ for a uniform distribution---a gap of only 1.2\%. The corresponding perplexity (effective number of attended stocks) is 278 out of 300. Figure~\ref{fig:heatmap} shows the time-averaged attention matrix sorted by Shenwan Level-1 industry: no block-diagonal structure is visible.

\begin{figure}[h]
\centering
\begin{minipage}{0.48\linewidth}
\centering
\includegraphics[width=\linewidth]{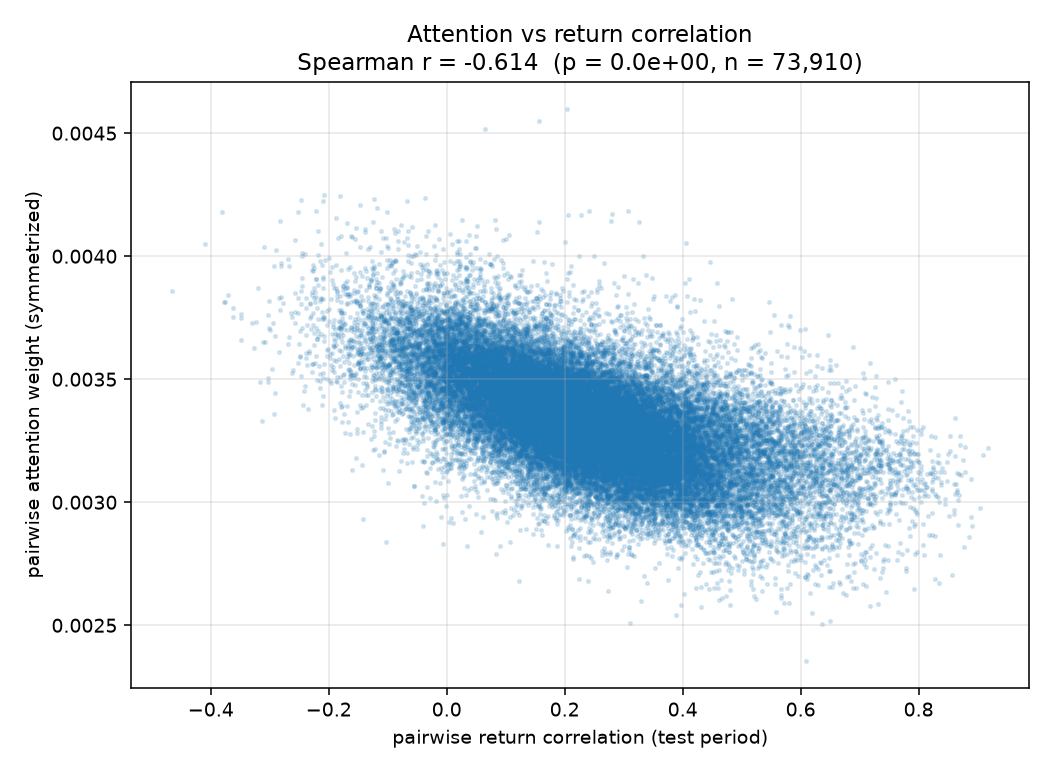}
\end{minipage}\hfill
\begin{minipage}{0.48\linewidth}
\centering
\includegraphics[width=\linewidth]{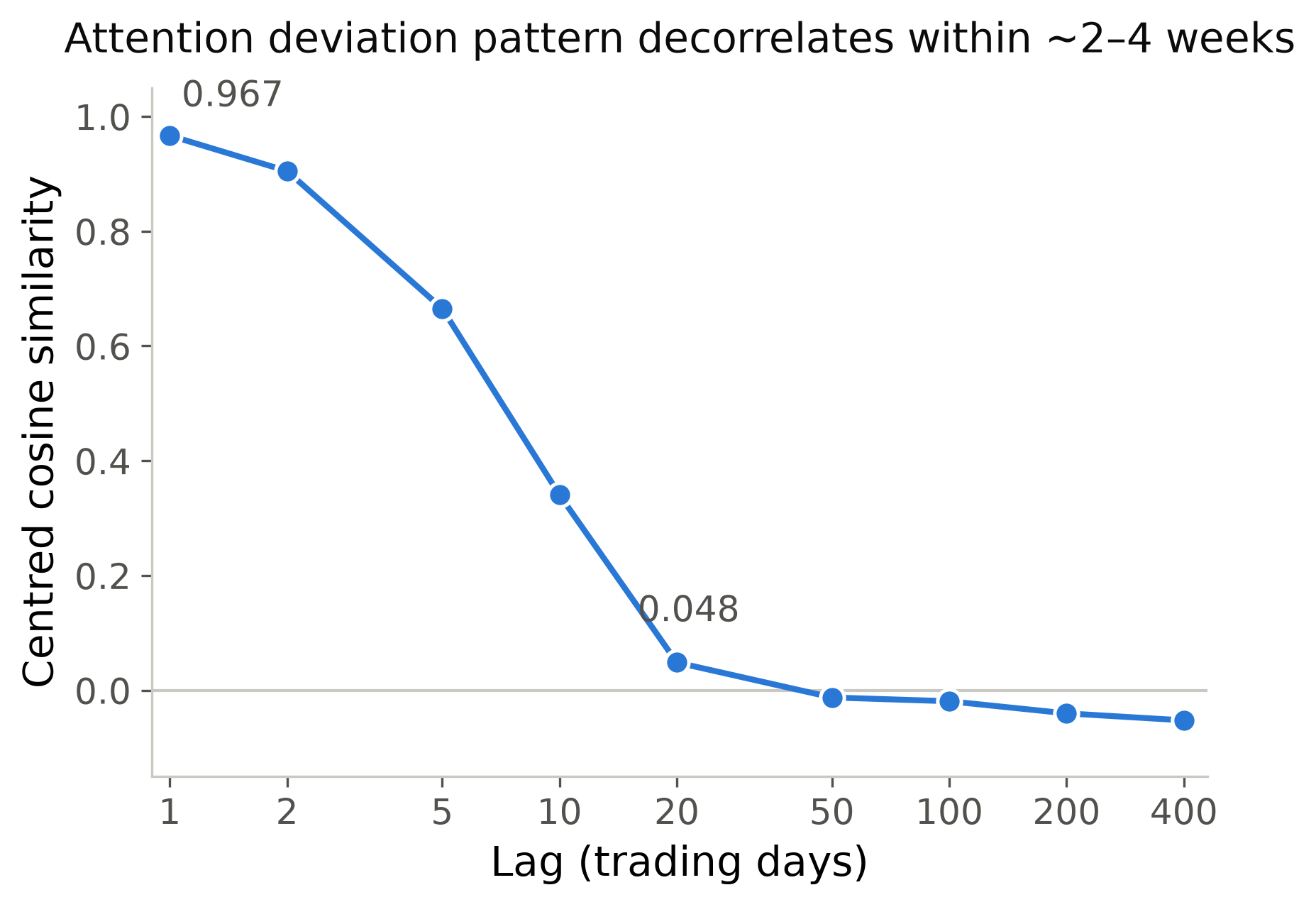}
\end{minipage}
\caption{\textbf{Left}: Pairwise attention weight vs.\ return correlation ($\rho = -0.614$; $n = 73{,}910$ pairs). \textbf{Right}: Centered cosine similarity vs.\ lag; full decorrelation in $\sim$20 trading days.}
\label{fig:scatter}
\label{fig:lagdecay}
\end{figure}

\paragraph{Anti-correlation with return similarity.} The Spearman correlation between pairwise attention weights and pairwise return correlations is $\rho = -0.614$ ($p \approx 0$, $n = 73{,}910$ valid pairs from a universe of 433 stocks; pairs require $\geq$60 overlapping return observations). Stocks that co-move receive \emph{less} attention. This association is robust to controls: on the 423 stocks (out of 433) with Shenwan L1 industry labels, the unconditional $\rho = -0.645$ (71,165 both-labeled pairs); a multiple regression controlling for same-industry membership, beta difference, and volatility difference yields a partial $\rho = -0.627$. Controls attenuate the association only marginally (Figure~\ref{fig:scatter}). Same-industry stock pairs receive significantly \emph{lower} attention than cross-industry pairs ($b = -0.011$, $p = 3 \times 10^{-4}$). Per-day Spearman $\rho$ is negative on all 619 test days (mean $-0.215$, std $0.071$). The per-day mean is substantially smaller than the pooled estimate because pooling across days absorbs shared cross-sectional structure that is constant within each day but varies across days.

\paragraph{Temporal dynamics.} Centered cosine similarity between attention matrices decays as: lag-1: 0.967, lag-5: 0.665, lag-20: 0.048 (Figure~\ref{fig:lagdecay}). The attention pattern fully reorganizes every 2--4 weeks. PCA of flattened daily deviation matrices shows PC1 explaining only 9\% of variance, indicating drift through high-dimensional space rather than oscillation between discrete modes. A permutation test confirms that market regime (up/flat/down days) does not drive attention changes ($p = 0.17$).

\paragraph{Preliminary interpretation.} By conventional entropy diagnostics, the attention is ``effectively uniform.'' As we show next, this interpretation is misleading---the 1.2\% deviation from uniformity carries all of the module's cross-sectional value.

\section{What Carries the Value?}
\label{sec:decomposition}

\subsection{Graph-Guided Variants All Fail}

If near-uniform attention wastes capacity on irrelevant stock pairs, explicit graph structure should help. We test three approaches---hard masking (GraphMask), soft biasing (GraphBias), and GCN replacement~\cite{gcn-kipf}---using both return-correlation and industry-based graphs. Table~\ref{tab:variants} shows that \emph{every} variant degrades performance. Hard masking to correlation-based neighbors produces IC \emph{below} complete module removal ($0.045 \pm 0.004$ vs.\ $0.050 \pm 0.001$ across 4 seeds; Table~\ref{tab:variants}), directly contradicting the intuition that attention should focus on ``related'' stocks.

\begin{table}[h]
\centering
\caption{Step~\ding{194} variants (CSI300). Multi-seed ($n=4$) for key configs; others seed 0. Ordering Original $>$ No Step~\ding{194} $>$ GraphMask holds across seeds (paired $t$: Original vs.\ GraphMask $p=0.040$).}
\label{tab:variants}
\begin{tabular}{lcccc}
\toprule
\textbf{Configuration} & \textbf{IC} & \textbf{Rank IC} & \textbf{ICIR} & \textbf{Seeds} \\
\midrule
Original MASTER & $0.057 \pm 0.008$ & $0.066 \pm 0.003$ & 0.386 & 4 \\
\midrule
GraphBias + Corr & 0.0497 & 0.0580 & 0.319 & 1 \\
GCNReplace + Corr & 0.0487 & 0.0570 & 0.374 & 1 \\
No Step~\ding{194} & $0.050 \pm 0.001$ & $0.052 \pm 0.002$ & 0.377 & 4 \\
GraphMask + Industry & 0.0437 & 0.0458 & 0.300 & 1 \\
GraphMask + Corr & $0.045 \pm 0.004$ & $0.049 \pm 0.002$ & 0.340 & 4 \\
\midrule
StaticAttn (oracle $\bar{A}$) & 0.0497 & 0.0595 & 0.382 & 1 \\
UniformAttn ($1/N$) & 0.0491 & 0.0514 & 0.368 & 1 \\
MeanPool & 0.0502 & 0.0536 & 0.377 & 1 \\
\bottomrule
\end{tabular}
\end{table}

\subsection{The Deviation Is Small but Essential}

To understand why near-uniform attention outperforms all alternatives, we decompose Step~\ding{194}'s output. For each stock $u$:
\begin{equation}
z_u = \underbrace{\frac{1}{N}\sum_v V_v}_{z_{\text{uniform}}} + \underbrace{\sum_v \left(\alpha_{uv} - \frac{1}{N}\right) V_v}_{z_{\text{deviation},u}}
\end{equation}
The uniform component $z_{\text{uniform}}$ is identical for all stocks (a rank-1 broadcast) and contributes 98.7\% of output energy. The deviation component $z_{\text{deviation},u}$ contributes only 1.3\% of energy but is the \emph{sole} source of cross-sectional variance---without it, Step~\ding{194}'s output carries no stock-specific information. We verify that value-vector heterogeneity does not affect this conclusion: the coefficient of variation of $\|V_v\|$ across stocks is 0.10, and the entropy of effective weights $\alpha_{uv}\|V_v\|$ differs from raw attention entropy by only 0.002 nats.

\subsection{Dynamic Recomputation Is Essential}

We freeze the attention to the time-averaged oracle matrix $\bar{A}$ (computed on the test set, giving the static pattern its best chance). StaticAttn recovers $\sim$47\% of Rank~IC value and only $\sim$9\% of IC value relative to the floor. Combined with the temporal dynamics analysis (Section~\ref{sec:diagnostic}), this establishes that value resides in daily recomputation, not in a stable backbone.

\section{Spectral Analysis and Nystr\"{o}m Approximation}
\label{sec:nystrom}

\subsection{The Deviation Matrix Is Low-Rank}

The preceding sections establish that Step~\ding{194}'s value lies in a near-uniform, dynamic, anti-correlation-seeking attention pattern. Sparsification fails because it destroys near-global support. We now ask: is there a \emph{different} axis of approximation?

We compute the SVD of the deviation matrix $D = A - \frac{1}{N}\mathbf{1}\mathbf{1}^\top$ averaged over test days. The effective rank is $\sim$65 (out of 300), and the top-10 singular values capture 96.5\% of $\|D\|_F^2$. This structure is robust: across 208 individual day$\times$head matrices, the median effective rank is 65 [IQR: 45--81] and the median top-10 energy share is 96.8\%.

\paragraph{Is low-rank structure learned or inherent?} We compute the deviation spectrum of 10 randomly initialized (untrained) MASTER models. The random-init deviation is \emph{also} low-rank (effective rank $59 \pm 7$, top-10 energy $95.6 \pm 0.9\%$), comparable to the trained model (rank $\sim$65, top-10 energy 96.5\%). Training modestly increases the effective rank and lowers entropy (5.49 vs.\ 5.68), but the low-rank property is present before any learning occurs. Thus, low-rank is inherent to softmax attention over $N$ items; what training contributes is the \emph{specific pattern} of deviations carrying predictive value. Nystr\"{o}m succeeds because softmax attention is architecturally guaranteed to remain low-rank, regardless of training.

This low-rank structure has a direct algorithmic implication: the Nystr\"{o}m method approximates the full attention matrix using $m \ll N$ landmark points, effectively performing a low-rank factorization. If the deviation matrix \emph{is} low-rank, Nystr\"{o}m should preserve its structure at $O(mN)$ cost.

\subsection{Nystr\"{o}m Matches Full Attention}

We replace Step~\ding{194}'s full attention with Nystr\"{o}m attention~\cite{nystromformer} using $m=32$ landmarks (10.7\% of $N$), selected uniformly at random each forward pass. The pseudo-inverse is computed via 6 Newton--Schulz iterations to avoid materializing any $N \times N$ matrix. All other components (FFN, LayerNorm, residual connections) are unchanged.

\begin{table}[h]
\centering
\caption{Nystr\"{o}m vs.\ full attention (CSI300, 5 seeds). TOST equivalence: Rank~IC within $\pm 0.005$ ($p=0.003$); IC within $\pm 0.008$ ($p=0.028$).}
\label{tab:nystrom}
\begin{tabular}{lcccc}
\toprule
& \textbf{IC} & \textbf{Rank IC} & \textbf{ICIR} & \textbf{Rank ICIR} \\
\midrule
Original ($O(N^2)$) & $0.058 \pm 0.007$ & $0.066 \pm 0.003$ & 0.386 & 0.429 \\
Nystr\"{o}m $m$=32 ($O(32N)$) & $0.059 \pm 0.002$ & $0.066 \pm 0.003$ & 0.389 & 0.416 \\
\midrule
\multicolumn{5}{l}{TOST ($\delta=0.005$): Rank~IC $p=0.003$ \checkmark\quad IC $p=0.099$} \\
\multicolumn{5}{l}{TOST ($\delta=0.008$): IC $p=0.028$ \checkmark\quad 90\% CI: IC $[-0.005, +0.007]$, Rank~IC $[-0.002, +0.001]$} \\
\bottomrule
\end{tabular}
\end{table}

The result (Table~\ref{tab:nystrom}) confirms the spectral prediction. A TOST equivalence test~\cite{tost} establishes that Nystr\"{o}m Rank~IC is equivalent to full attention within $\pm 0.005$ ($p = 0.003$); IC equivalence holds at the $\pm 0.008$ margin ($p = 0.028$) but not at $\pm 0.005$ ($p = 0.099$), reflecting higher IC variance across seeds. Nystr\"{o}m also exhibits notably lower seed-to-seed IC variance ($0.002$ vs.\ $0.007$). Landmark-induced stochasticity is negligible (eval jitter $< 10^{-4}$).

\subsection{Why Low-Rank Works but Sparse Does Not}

Table~\ref{tab:sweep} compares the full landscape of efficient alternatives. Two patterns emerge. First, low-rank methods dominate sparse methods: Nystr\"{o}m~($m$=32) matches the original on Rank~IC while TopK~($K$=16) retains at most 50\% of either metric (IC equivalence across seeds is established in Table~\ref{tab:nystrom}). Second, \emph{more} landmarks do not help: $m$=64 performs worse than $m$=32, a phenomenon we analyze in detail in Section~\ref{sec:msweep}. Landmark-based approximation (Nystr\"{o}m) also substantially outperforms random-feature approximation (Performer) at the same $m$.

\begin{table}[h]
\centering
\caption{Efficient attention sweep (CSI300). Multi-seed for the main comparisons; exploratory rows are seed~0 only. Low-rank dominates sparse; landmark count is non-monotonic (see \S\ref{sec:msweep}).}
\label{tab:sweep}
\begin{tabular}{lcccc}
\toprule
\textbf{Configuration} & \textbf{IC} & \textbf{Rank IC} & \textbf{Complexity} & \textbf{Seeds} \\
\midrule
Original & $0.058 \pm 0.007$ & $0.066 \pm 0.003$ & $O(N^2)$ & 5 \\
Nystr\"{o}m $m$=32 & $0.059 \pm 0.002$ & $0.066 \pm 0.003$ & $O(32N)$ & 5 \\
DeviationAttn & $0.052 \pm 0.005$ & $0.065 \pm 0.004$ & $O(N^2)$* & 5 \\
TopK $K$=16 & $0.052 \pm 0.004$ & $0.064 \pm 0.004$ & $O(16N)$ & 5 \\
No Step~\ding{194} (floor) & $0.050 \pm 0.001$ & $0.052 \pm 0.002$ & --- & 4 \\
\midrule
\multicolumn{5}{l}{\emph{Seed 0 only (exploratory):}} \\
Nystr\"{o}m $m$=64 & 0.0554 & 0.0603 & $O(64N)$ & 1 \\
Performer $m$=32 & 0.0540 & 0.0577 & $O(32N)$ & 1 \\
UniformAttn & 0.0491 & 0.0514 & $O(N^2)$ & 1 \\
\bottomrule
\multicolumn{5}{l}{\small *Requires full $N^2$ softmax before thresholding.} \\
\end{tabular}
\end{table}

The mechanistic explanation follows from the near-uniform attention structure. In a distribution where most weights are $\approx 1/N$, the top-$K$ entries are themselves near-equal (perplexity 15.9/16 at $K$=16). Subtracting a constant from near-equal values cannot create meaningful differentiation---sharpening requires support spanning a range around $1/N$, which small $K$ cannot provide. This is a \emph{structural} prediction: all small-$K$ weighted variants should fail identically, and our 2$\times$2 factorial (varying support size $\times$ weighting rule) confirms this.

\subsection{Landmark Count: A Non-Monotonic Profile}
\label{sec:msweep}

\begin{figure}[h]
\centering
\includegraphics[width=0.9\linewidth]{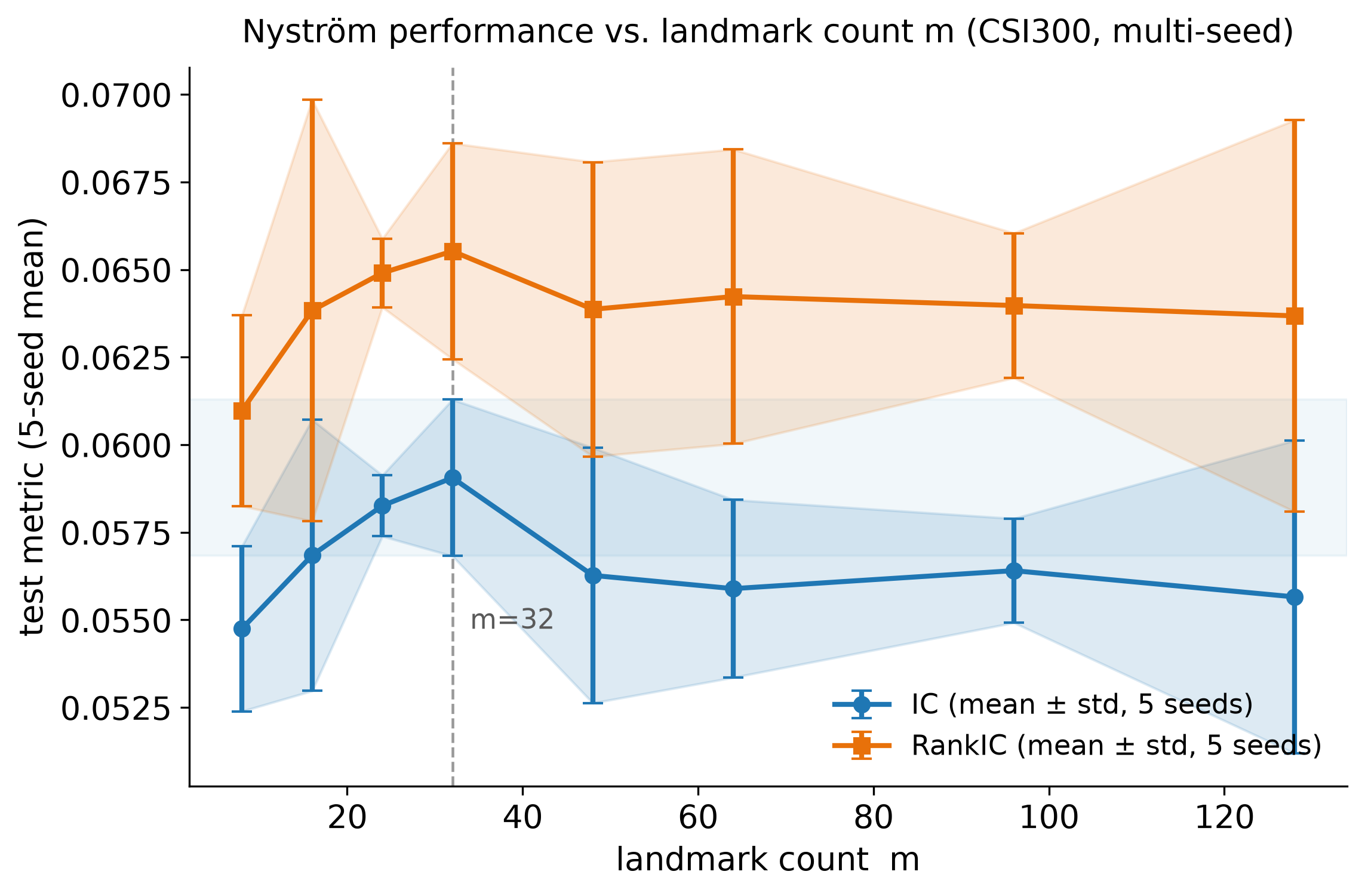}
\caption{Nystr\"{o}m performance vs.\ landmark count $m$ (CSI300, 5 seeds each, $\pm 1$ std). Peak at $m=32$; non-monotonic profile reflects bias-variance trade-off in pseudo-inverse computation.}
\label{fig:msweep}
\end{figure}

A dense sweep of $m \in \{8, 16, 24, 32, 48, 64, 96, 128\}$ with 5 seeds each reveals a non-monotonic relationship between landmark count and performance (Figure~\ref{fig:msweep}). Performance peaks at $m=32$ (IC $0.059 \pm 0.002$, Rank~IC $0.066 \pm 0.003$), dips at $m=48$ ($1.2\sigma$ below $m=32$), and remains below the peak at $m=64$--128. This non-monotonic profile, confirmed across seeds, resolves an apparent contradiction: $m=64$ underperforms $m=32$ despite the deviation matrix having effective rank $\sim$65. The explanation is that $m$ landmarks in Nystr\"{o}m do not correspond one-to-one to SVD components; over-sizing the landmark set worsens the conditioning of the pseudo-inverse, increasing approximation variance faster than it reduces bias.

\subsection{Dynamic Sampling Is Essential, Not Just Low-Rank Structure}

To disentangle whether Nystr\"{o}m's success derives from its low-rank structure or its dynamic (per-forward-pass) landmark resampling, we test a \textbf{LearnedLowRank} baseline: each stock receives a learnable embedding $e_i \in \mathbb{R}^{32}$, and attention scores are computed as $a_{ij} = e_i^\top e_j / \sqrt{32}$ (static, trained end-to-end). This provides a rank-32 attention pattern that is \emph{optimized for the task} but \emph{fixed across days}.

\begin{table}[h]
\centering
\caption{Dynamic vs.\ static, crossed with full-rank vs.\ low-rank (CSI300, seed~0; 5-seed means for Original and Nystr\"{o}m are in Table~\ref{tab:nystrom}). Dynamic routing is essential; low-rank is viable only when dynamic.}
\label{tab:dynamic_lowrank}
\begin{tabular}{lcccc}
\toprule
& \textbf{IC} & \textbf{Rank IC} & \textbf{Dynamic?} & \textbf{Rank} \\
\midrule
Full attention & 0.0646 & 0.0685 & Yes & Full \\
Nystr\"{o}m $m$=32 & 0.0607 & 0.0685 & Yes (random) & Low \\
StaticAttn (oracle) & 0.0497 & 0.0595 & No (frozen) & Full \\
LearnedLowRank $r$=32 & 0.0478 & 0.0500 & No (learned) & Low \\
No Step~\ding{194} & 0.0482 & 0.0514 & --- & --- \\
\bottomrule
\end{tabular}
\end{table}

Table~\ref{tab:dynamic_lowrank} reveals a clear interaction: low-rank approximation preserves performance only when combined with dynamic, data-dependent routing (Nystr\"{o}m). A static low-rank attention, even when optimized end-to-end, collapses to the no-attention floor ($\text{IC} = 0.0478 \approx 0.0482$). This establishes that Nystr\"{o}m's success is not merely a consequence of low-rank structure but requires the QK-driven, day-specific attention computation that adapts to each day's cross-sectional configuration.

\section{Cross-Stock Aggregation at Scale}
\label{sec:scale}

The Nystr\"{o}m result enables, in principle, scaling cross-stock attention to large universes at $O(mN)$ cost. We test scaling along two axes: first within the same MASTER codebase at $N=800$ (CSI800), then with adapted architectures at $N \approx 3{,}500$.

\subsection{CSI800: Same Codebase, Larger Universe}

Using the same MASTER architecture, features, and training protocol as CSI300---only $N$ changes---we compare Original vs.\ Nystr\"{o}m at $N \approx 800$ (10 seeds each).

\begin{table}[h]
\centering
\caption{Nystr\"{o}m at CSI800 ($N \approx 800$, 10 seeds). Equivalence certified within $\pm 0.005$ on both metrics.}
\label{tab:csi800}
\begin{tabular}{lcc}
\toprule
& \textbf{Original ($n$=10)} & \textbf{Nyst $m$=32 ($n$=10)} \\
\midrule
IC & $0.047 \pm 0.004$ & $0.045 \pm 0.004$ \\
Rank IC & $0.059 \pm 0.004$ & $0.059 \pm 0.006$ \\
\midrule
\multicolumn{3}{l}{TOST ($\delta=0.005$): IC $p=0.038$ \checkmark\quad RankIC $p=0.034$ \checkmark} \\
\multicolumn{3}{l}{90\% CI: IC $[-0.005, +0.001]$\quad RankIC $[-0.004, +0.004]$} \\
\bottomrule
\end{tabular}
\end{table}

At $N=800$, Nystr\"{o}m with $m=32$ (4\% coverage) is certified equivalent to full attention within $\pm 0.005$ on both IC ($p=0.038$) and Rank~IC ($p=0.034$). Importantly, increasing landmarks to $m=80$ (10\%, matching CSI300's ratio) does \emph{not} improve performance---the optimal $m \approx 32$ is stable across scale. An initial 5-seed analysis suggested a possible gap; doubling to 10 seeds revealed this as a statistical-power artifact, not a real degradation.

\subsection{$N \approx 3{,}500$: Adapted Architecture}

We further test at $N \approx 3{,}500$ using the full A-share market with adapted architectures (17 OHLCV-derived features, $d_{\text{model}}=64$, no market gating). Table~\ref{tab:scale} reports results across 4 seeds.

\begin{table}[htbp]
\centering
\caption{Large-scale validation ($N \approx 3{,}500$, 4 seeds). No cross-stock module significantly outperforms PureLSTM.}
\label{tab:scale}
\begin{tabular}{lcccc}
\toprule
& \textbf{IC} & \textbf{Rank IC} & \textbf{Sharpe} \\
\midrule
PureLSTM & $0.043 \pm 0.002$ & $0.056 \pm 0.003$ & $5.6 \pm 0.5$ \\
LSTM + GCN & $0.034 \pm 0.008$ & $0.063 \pm 0.009$ & $4.1 \pm 1.5$ \\
MASTER + Nyst32 & $0.037 \pm 0.008$ & $0.050 \pm 0.016$ & $5.6 \pm 1.1$ \\
\bottomrule
\end{tabular}
\end{table}

At this scale, no cross-stock module significantly outperforms PureLSTM on either IC or Rank~IC (paired $t$-tests: MASTER+Nyst32 vs.\ PureLSTM IC $p=0.27$, Rank~IC $p=0.53$; $n=4$). An initial single-seed analysis (seed 42) had suggested a consistent IC$\uparrow$/Rank~IC$\downarrow$ trade-off across all module types, but this pattern did not survive multi-seed validation---underscoring the necessity of multi-seed evaluation in financial ML. The high seed-to-seed variance at $N \approx 3{,}500$ (e.g., MASTER+Nyst32 IC std $= 0.008$, vs.\ $0.002$ at CSI300) suggests that the noisier, more heterogeneous full-market cross-section is substantially harder to model, and that Nystr\"{o}m's benefits---established at $N = 300$ and $N = 800$ within MASTER's native pipeline---do not trivially transfer to adapted architectures at larger scale.

\section{Efficiency}
\label{sec:efficiency}

\begin{figure}[h]
\centering
\includegraphics[width=0.95\linewidth]{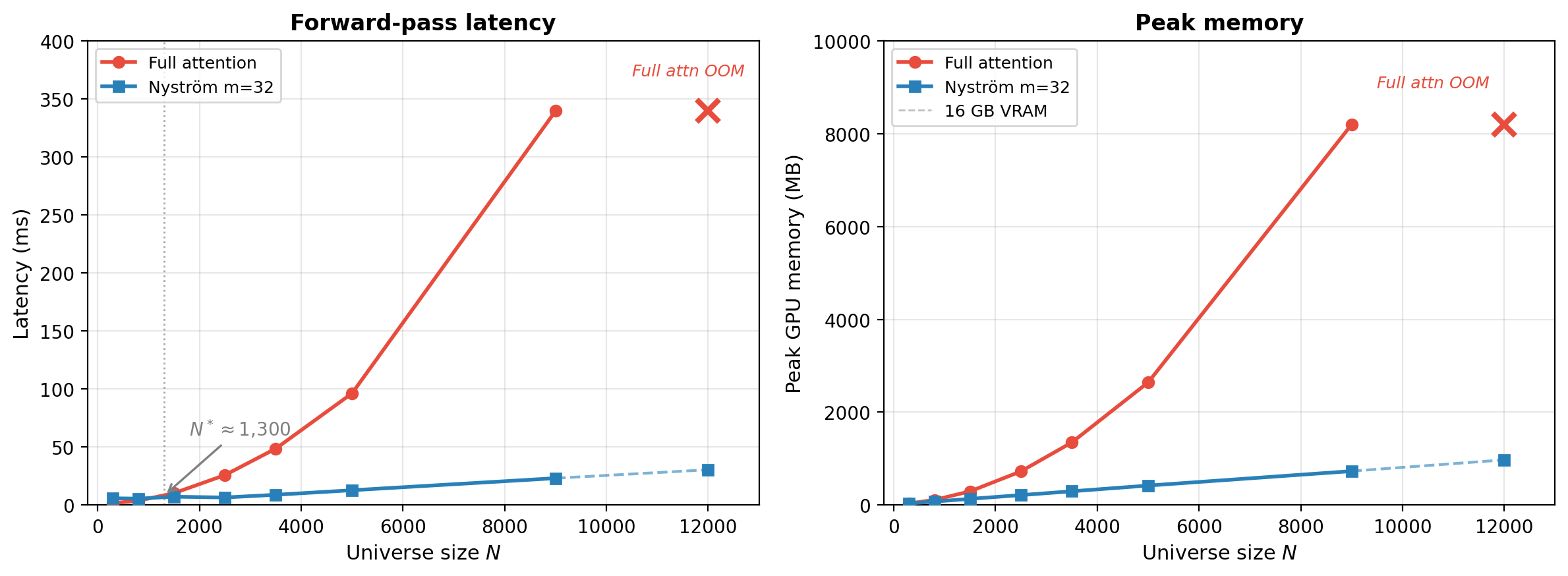}
\caption{Forward-pass latency and memory vs.\ $N$ on a Tesla T4 (16\,GB), extended to $N = 12{,}000$. Full attention OOMs at $N = 12{,}000$; Nystr\"{o}m scales linearly and uses 974\,MB there. Latency crossover at $N^* \approx 1{,}300$.}
\label{fig:efficiency}
\end{figure}

The equivalence results establish that Nystr\"{o}m preserves predictive quality. We now quantify the computational savings via forward-pass benchmarks on a Tesla T4 (16\,GB, CUDA 12.8), using the deployed Step~\ding{194} configuration ($d_{\text{model}} = 256$, $n_{\text{head}} = 2$, $T = 8$), with 50 timed passes after 10 warmup iterations (Table~\ref{tab:efficiency}, Figure~\ref{fig:efficiency}).

\begin{table}[h]
\centering
\caption{Forward-pass efficiency: full attention vs.\ Nystr\"{o}m ($m = 32$). Median latency (ms) and peak GPU memory (MB) on a Tesla T4 (16\,GB). $d_{\text{model}} = 256$, $n_{\text{head}} = 2$, $T = 8$.}
\label{tab:efficiency}
\begin{tabular}{rcccccc}
\toprule
& \multicolumn{2}{c}{\textbf{Latency (ms)}} & & \multicolumn{2}{c}{\textbf{Memory (MB)}} & \\
\cmidrule{2-3} \cmidrule{5-6}
$N$ & Full & Nyst & Speedup & Full & Nyst & Saving \\
\midrule
300 & 1.65 & 5.90 & 0.3$\times$ & 37 & 36 & 1.0$\times$ \\
800 & 3.80 & 5.59 & 0.7$\times$ & 112 & 77 & 1.5$\times$ \\
1{,}500 & 10.30 & 7.24 & 1.4$\times$ & 297 & 135 & 2.2$\times$ \\
2{,}500 & 25.87 & 6.57 & 3.9$\times$ & 727 & 216 & 3.4$\times$ \\
3{,}500 & 48.66 & 8.92 & 5.5$\times$ & 1{,}350 & 298 & 4.5$\times$ \\
5{,}000 & 96.30 & 12.82 & 7.5$\times$ & 2{,}642 & 421 & 6.3$\times$ \\
9{,}000 & 339.5 & 23.17 & 14.7$\times$ & 8{,}194 & 732 & 11.2$\times$ \\
12{,}000 & OOM & 30.46 & --- & OOM & 974 & --- \\
\bottomrule
\end{tabular}
\end{table}

At $N = 300$, Nystr\"{o}m is \emph{slower} (5.90 vs.\ 1.65\,ms): the Newton--Schulz pseudo-inverse imposes a fixed overhead of $\sim$4\,ms that is independent of $N$.\footnote{An iteration-count ablation confirms this: at $N = 300$, Nystr\"{o}m latency rises from 2.29\,ms (0 iterations) to 5.95\,ms (6 iterations), i.e.\ $\sim$0.6\,ms per iteration. At $N = 3{,}500$ the same 6 iterations add only 0.4\,ms (9.07 $\to$ 9.48\,ms), because the $O(mN)$ attention computation dominates.} The latency crossover occurs at $N^* \approx 1{,}300$; beyond this, Nystr\"{o}m's $O(mN)$ scaling wins decisively---$5.5\times$ at $N = 3{,}500$ and $14.7\times$ at $N = 9{,}000$. The memory scaling is unambiguous: from $N = 300$ to $9{,}000$ (a $30\times$ increase), full-attention memory grows $\times 219$ (consistent with $\propto N^2$) while Nystr\"{o}m grows $\times 20$ (consistent with $\propto N$). Full attention exhausts the 16\,GB budget at $N = 12{,}000$, where Nystr\"{o}m still requires only 974\,MB.

\paragraph{Training overhead.} Nystr\"{o}m training is $\sim$2$\times$ slower per epoch due to Newton--Schulz iterations (6 sequential matrix multiplications). Training is a one-time offline cost; daily inference for portfolio rebalancing is the recurring cost that scales with $N$. At CSI300 we do not claim wall-clock savings---full attention is in fact faster below $N^* \approx 1{,}300$. The value proposition is at scale: a $5$--$15\times$ latency reduction with proportional memory savings for $N > 3{,}000$, at equivalent predictive quality.

\section{SVD Factor Interpretation}
\label{sec:svd}

The spectral analysis reveals \emph{that} the deviation is low-rank; we now ask \emph{what} each component encodes. Mapping the top SVD components of $D$ to financial characteristics reveals that attention implicitly implements a multi-factor risk model (Table~\ref{tab:factors}).

\begin{table}[h]
\centering
\caption{Top SVD components of the deviation matrix mapped to financial risk factors. Six components (91\% of energy) have interpretable financial meaning; PCs 6 and 8--10 do not.}
\label{tab:factors}
\begin{tabular}{cccccl}
\toprule
\textbf{PC} & \textbf{Energy} & \textbf{Beta $\rho$} & \textbf{Vol $\rho$} & $\eta^2$ & \textbf{Interpretation} \\
\midrule
1 & 79.5\% & +0.11 & +0.48 & 0.12 & Volatility / market mode \\
2 & 3.6\% & $-$0.04 & $-$0.05 & 0.14 & Momentum (pharma vs.\ power equip.) \\
3 & 3.4\% & +0.02 & $-$0.02 & 0.18 & Old vs.\ new economy \\
4 & 2.0\% & $-$0.53 & $-$0.69 & 0.48 & Defensive vs.\ cyclical \\
5 & 1.4\% & $-$0.25 & $-$0.23 & 0.22 & Liquidity / turnover \\
7 & 1.1\% & $-$0.05 & $-$0.24 & 0.21 & Reversal \\
\midrule
6,8--10 & 3.5\% & --- & --- & --- & Not interpretable \\
\bottomrule
\end{tabular}
\end{table}

The dominant mode (PC1, 79.5\%) tracks cross-sectional volatility dispersion---high-volatility stocks load on one end, low-volatility on the other. This single mode explains both the anti-correlation finding (high-vol stocks co-move and cluster on the same side of PC1, receiving less differentiating attention) and Nystr\"{o}m's success (a structure dominated by one mode is trivially captured by few landmarks).

PC4 provides the cleanest industry signal ($\eta^2 = 0.48$): banks and construction (defensive, low-beta) vs.\ autos and power equipment (cyclical, high-beta). PCs 2, 3, 5, and 7 correspond to momentum, sector rotation (old vs.\ new economy), liquidity, and reversal---canonical factors in the quantitative finance literature.

MASTER's inter-stock attention thus functions as an \emph{implicit 6-factor risk model}: it routes information along dimensions that align with well-known financial risk factors, despite never being explicitly trained on factor labels.

\section{Discussion}
\label{sec:discussion}

\paragraph{Entropy diagnostics are insufficient.} Perplexity 278/300 would lead most practitioners to conclude that attention is ``effectively uniform''---an interpretation our spectral analysis contradicts. Entropy is a bulk statistic that cannot distinguish every weight equaling $1/N$ from most weights near $1/N$ with a structured low-rank deviation. We recommend that attention analysis always include functional ablations (uniform-forcing, oracle-static) alongside distributional summaries.

\paragraph{Complementarity, not similarity.} The robust negative association between attention and return correlation ($\rho = -0.614$ unconditionally, $-0.627$ after controlling for industry, beta, and volatility; negative on all 619 test days) suggests that inter-stock attention implements information \emph{diversification}: each stock attends preferentially to stocks with different price dynamics. This is the opposite of MASTER's stated ``correlation mining'' interpretation and is consistent with the failure of correlation-based graph masking.

\paragraph{IC vs.\ Rank~IC dissociation.} A recurring pattern across our CSI300 experiments is that IC and Rank~IC respond to different mechanisms. IC scales with the \emph{breadth} of aggregation (neighbor count), while Rank~IC depends on weight \emph{sharpness}. Tercile analysis by return magnitude reveals that predictive signal is overwhelmingly concentrated in extreme movers: the top tercile by $|\text{return}|$ carries $\sim$77\% of IC covariance at CSI300, with 3--4$\times$ the IC of near-zero stocks. This dissociation has practical implications: portfolio strategies focused on return magnitude may prefer cross-stock modules, while pure ranking strategies may be better served by per-stock models. However, at $N \approx 3{,}500$, neither advantage survives multi-seed validation (Table~\ref{tab:scale}), suggesting that this dissociation is most relevant within MASTER's native architecture and scale.

\paragraph{Related work.} Our finding that near-uniform attention carries predictive value echoes work in NLP showing that fixed or random attention patterns can match learned attention~\cite{jain-wallace,synthesizer}. We extend this to a financial domain and add a spectral explanation. StockMixer~\cite{stockmixer}, from the same research group as MASTER, achieves comparable performance with a static MLP-based cross-stock mixing layer, consistent with our finding that the attention pattern's effective mechanism is near-global redistribution rather than selective routing.

\paragraph{Economic significance.} All Sharpe ratios reported are frictionless (no transaction costs, slippage, or market impact). In the A-share market, where daily turnover is high and short-selling is restricted, the IC differences of 0.005--0.01 we observe are modest but practically relevant for institutional portfolios. Practitioners should note that IC and Rank~IC serve different downstream tasks: return-focused strategies (e.g., long-short with continuous position sizing) benefit from IC improvements, while pure ranking strategies (e.g., top-$K$ selection) depend on Rank~IC.

\paragraph{Preprocessing sensitivity.} Under improved feature normalization (per-day cross-sectional z-score, beyond Qlib's default global robust z-score), full attention benefits more than Nystr\"{o}m: Original IC improves by $+0.005$ with 5.8$\times$ lower seed variance, while Nystr\"{o}m IC is unchanged. The TOST Rank~IC equivalence weakens from $p=0.003$ to $p=0.120$, suggesting the equivalence margin is partially contingent on input conditioning. Better-conditioned inputs allow full $O(N^2)$ attention to extract finer cross-sectional structure that the low-rank approximation misses. This represents an honest boundary of our core claim: Nystr\"{o}m equivalence is robust under standard preprocessing but narrows under optimized preprocessing.

\paragraph{Limitations.} The CSI300 decomposition results rely on seed~0 for most ablations; the key Nystr\"{o}m claim is validated across 5 seeds (CSI300) and 10 seeds (CSI800). The large-scale experiments use an adapted MASTER pipeline (17 features, $d_{\text{model}}=64$, no market gating) rather than the full original architecture, limiting direct comparability with the CSI300 analysis. Our findings are established on a single model (MASTER) in a single market (Chinese A-shares); whether the low-rank deviation structure is a general property of cross-stock attention or specific to MASTER requires verification on additional architectures and markets.

\section{Conclusion}

We have systematically decomposed MASTER's inter-stock attention module and established that its value derives from a mechanism fundamentally different from what its design implies. Rather than learning interpretable stock relationships, the module implements a dynamic, near-global redistribution whose deviation from uniformity is low-rank and compressible. This low-rank structure---inherent to softmax attention, not learned---explains why Nystr\"{o}m approximation matches full attention on ranking metrics at $O(mN)$ cost (TOST-certified at $N=300$ and $N=800$) while sparsification consistently fails. At $N \approx 3{,}500$ with adapted architectures, cross-stock modules do not significantly outperform per-stock baselines, indicating that the benefits do not trivially transfer across scales and architectures---a finding that itself required multi-seed validation to establish. Future work may explore cross-stock mechanisms that enhance differentiation rather than smooth it, as well as testing Nystr\"{o}m within MASTER's full pipeline at larger scales.

\bibliographystyle{plain}

\end{document}